\documentclass{neus2025}

\usepackage{times}
\usepackage{simplemacros}
\usepackage{mathtools}
\usepackage{graphicx} 
\usepackage{subcaption}
\usepackage{float}

\providecommand\itembox{}
\providecommand\subfiggraphicswidth{}
\newcommand{\Sone}{\mathcal{S}_1}
\newcommand{\Stwo}{\mathcal{S}_2}
\newcommand{\Dthree}{\mathcal{D}^{3D}}
\newcommand{\Dthreebar}{\Bar{\mathcal{D}}^{3D}}
\newcommand{\Dtwo}{\mathcal{D}^{2D}}

\title[Neuro-Symbolic Paradigm for Perception]{Assured Autonomy with Neuro-Symbolic Perception}

\coltauthor{
\Name{R. Spencer Hallyburton} \Email{spencer.hallyburton@duke.edu}\\
\Name{Miroslav Pajic} \Email{miroslav.pajic@duke.edu}\\
\addr Duke University
}

\begin{document}

\maketitle


\begin{abstract}
Many state-of-the-art AI models deployed in cyber-physical systems (CPS), while highly accurate, are simply pattern-matchers.~With limited security guarantees, there are concerns for their reliability in safety-critical and contested domains. To advance assured AI, we advocate for a paradigm shift that imbues data-driven perception models with symbolic structure, inspired by a human's ability to reason over low-level features and high-level context. We propose a neuro-symbolic paradigm for perception (NeuSPaPer) and illustrate how joint object detection and scene graph generation (SGG) yields deep scene understanding.~Powered by foundation models for offline knowledge extraction and specialized SGG algorithms for real-time deployment, we design a framework leveraging structured relational graphs that ensures the integrity of situational awareness in autonomy. Using physics-based simulators and real-world datasets, we demonstrate how SGG bridges the gap between low-level sensor perception and high-level reasoning, establishing a foundation for resilient, context-aware AI and advancing trusted autonomy in CPS.
\end{abstract}
\section{Introduction}

Over the past decade, AI research has been primarily focused on optimizing black-box models with vast domain-specific training data. While benchmark performance has improved, the effort spent tuning traditional models has not led to significant assuredness guarantees. It is well known that even minor perturbations to the input data of AI models, both natural and adversarial, have led to high-profile unintended and sometimes catastrophic failures (e.g., from~\cite{eykholt2018robust}, \cite{finlayson2019adversarial}), raising concerns about AI's reliability in safety-critical cyber-physical systems (CPS). 

Defensive techniques such as adversarial training \cite{shafahi2020universal}, distillation \cite{papernot2016distillation}, and ensembling \cite{jia2019certified} contend to secure models. However, adaptive adversaries consistently overcome such defenses \cite{carlini2017adversarial}, suggesting that deep neural networks (DNNs) are statistical pattern-matchers rather than true high-level reasoners. Current security analyses remain incomplete, focusing largely on structured noise like $L_p$ norm perturbations that fail to capture real-world adversary complexities. Examining recent attacks on multi-sensor fusion, such as the frustum attack \cite{hallyburton2022security}, we argue that traditional DNN architectures face innate and unavoidable vulnerabilities to attacks that alter a scene's semantic structure.

Achieving robust perception necessitates moving beyond reactive defenses to existing architectures and integrating structured reasoning with statistical learning.~In contrast to DNNs, human perception seamlessly integrates low-level feature recognition with high-level contextual and commonsense reasoning, enabling us to interpret ambiguous, noisy, or incomplete data because of an ability to infer object relationships, detect inconsistencies, and reason about cause-and-effect interactions in a scene. Given fundamental vulnerabilities of existing DNNs, we advocate for an incorporation of symbolic reasoning into perception models to enhance reliability and robustness.

In particular, we propose a paradigm shift in sensor fusion from vulnerable pattern-matching DNNs to logically-grounded reasoning algorithms that combine neural and symbolic components. Transcending black-box algorithms, a neuro-symbolic approach allows for incorporating logical constraints and commonsense knowledge -- an approach that promises to enhance robustness in safety-critical applications such as autonomous driving (AD) and unmanned aerial vehicles~(UAVs).

Our neuro-symbolic approach to sensor fusion commences with a joint detection and graph generation step leveraging advancements in scene graph generation (SGG), a promising backbone for grounding black-box inference in high level logical relationships. SGG algorithms build graphical representations of scenes by identifying objects (nodes) and illuminating salient interactions between them (edges). Applied to perception, SGG can enrich situational awareness with \emph{contextualized high- and low-level concepts} that traditional object detectors are ill-suited to discern.

To secure single- and multi-sensor fusion with SGG requires the design of specialized integrity algorithms for scene-graph-based anomaly detection. We propose a two-stage integrity framework where graphs from each sensor are first evaluated against physics-based knowledge bases to ensure compliance with commonsense understanding (per-sensor). Graphs from all sensors are then sent to a multi-sensor graph consistency evaluator that considers the compatibility of nodes and edges across graphs (cross-sensor). Insights from graph-based integrity are used to flag anomalous inference results and weight information updates in a downstream graph-informed sensor fusion step. 

In this early work, we present feasibility case studies to illustrate the promising potential of neuro-symbolic sensor fusion.~We walk through case studies based on both real-world datasets (nuScenes,~\cite{caesar2020nuscenes}) and physics-based simulators (CARLA,~\cite{dosovitskiy2017carla}). In the camera domain, we use a foundation model to predict graphs from RGB camera images. In the LiDAR domain, we use a rule-based approach because the data already fully resolve 3D Cartesian space. Object detections and graphs are then compared to illuminate any inconsistencies between them. We show that even when considering challenging attacks such as 
the \emph{frustum attacks} against the LiDAR sensor, we can easily identify incompatible scene graphs between the camera~and~LiDAR. 
To the best of our knowledge, this is the first single-platform approach demonstrated to detect such attacks, and results suggest the potential for neuro-symbolic methods to significantly improve security guarantees for  perception in CPS. These findings motivate planned future research in designing full-stack neuro-symbolic perception and integrity.

\vspace{4pt}
\noindent \textbf{Contributions.} In summary, the contributions of this work include:
\begin{itemize}
    \setlength\itemsep{-4pt}
    \item \textbf{Vulnerability analysis:} demonstration of the vulnerability of DNN models to stealthy, undetectable attacks on sensing that alter the semantic structure of the scene.
    \item \textbf{Neuro-symbolic perception:} Design of a novel neuro-symbolic paradigm for perception jointly performing detection, classification, and graph-building from sensor data. 
    \item \textbf{Neuro-symbolic integrity:} Architecting of neuro-symbolic integrity to reason over per-sensor and cross-sensor consistency against commonsense and physics-informed knowledge.
    \item \textbf{Feasibility study:} Case studies in real-world and simulated datasets demonstrating first single-platform detection of challenging attacks, previously thought to be stealthy. 
\end{itemize}
\section{Sensor Fusion in Autonomy}

\paragraph{Perception.}

DNNs are the state-of-the-art in object detection. Widely used algorithms/architectures for images (img) include convolutional neural networks (CNNs), \textbf{Faster R-CNN}~\cite{ren2016faster} and \textbf{YOLO}~\cite{redmon2016you}, while for point clouds (pc) analysis, \textbf{PointPillars}~\cite{lang2019pointpillars} and \textbf{PV-RCNN}~\cite{shi2020pv} are used. Recent \textbf{transformer-based detectors}, such as \textbf{DETR}~\cite{carion2020detr}, offer an end-to-end approach minimizing the use of hand-crafted heuristics. 

\paragraph{Multi-sensor fusion.}

Fusing data improves observability, robustness, and attack resilience. Multi-sensor fusion typically occurs at the semantic level~\cite{durrant2016multisensor}. Such approaches are used in AD~\cite{baresi2023architecting} and UAVs~\cite{fei2023uav}. Appendix~\ref{appendix:fusion} formalizes the semantic-level multi-sensor fusion problem.
\section{Vulnerability of Perception \& Sensor Fusion} \label{sec:vulnerability}

In this section, we discuss how attacks on perception that alter the semantic structure of a scene are indefensible by reactive hardening of algorithms due to the equivariance of DNNs. We illustrate that this is of significant concern showing a stealthy \emph{frustum attack} with high degree of attack success. 

\subsection{Equivariance Vulnerability}

In response to DNNs' vulnerability to out of distribution attacks (e.g.,~\(L_p\) norm), certified robustness techniques including sub-sampling/ensembling~\cite{jia2019certified} were proposed, yielding security guarantees. However, despite the effectiveness of certified robustness against \( L_p \) attacks, they fail to defend many real-world attacks that fundamentally alter the semantic structure of the scene.

Specifically, point-based (e.g., LiDAR) DNNs are designed so that features derived on a collection of points are invariant to spatial translations (see~\cite{bronstein2021geometric} on equivariance). Thus, no amount of sub-sampling or ensembling can mitigate \emph{translation attacks} because the original features are retained when points from a legitimate object are (adversarially) moved to a new location. Spoofing attacks where points are injected patterning a legitimate object yield similar outcomes. While equivariance enhances model accuracy and generalization, it unfortunately introduces a vulnerability that adversaries can exploit, rendering certified robustness techniques ineffective.

\subsection{Stealthy Attacks On Sensor Fusion} \label{sec:vulnerability-case}

\begin{figure}
    \centering
    \includegraphics[width=\linewidth]{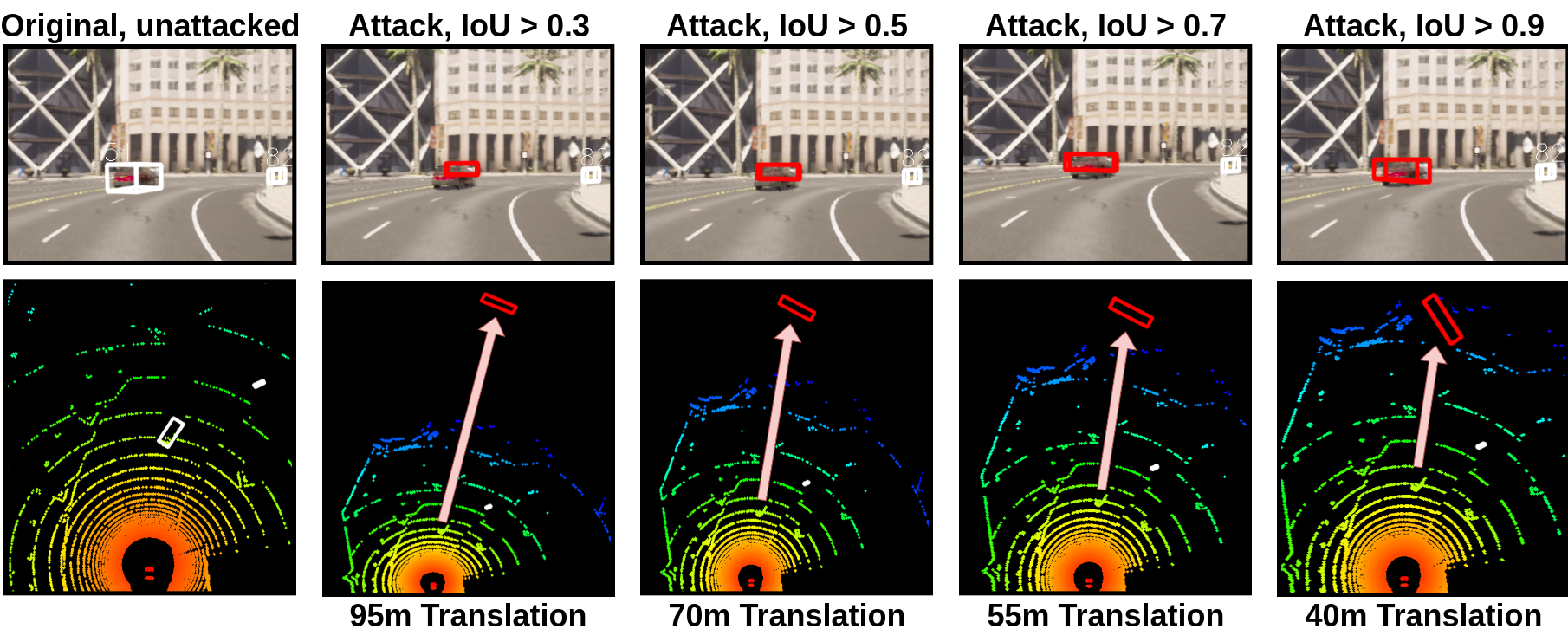}
    \vspace{-16pt}
    \caption{Attacker can alter the semantic understanding of the scene while being stealthy to multi-sensor fusion. Translating existing 3D objects (denoted with white box) backwards or forwards (resulting in the detected 'moved' red boxes) from ego maintains consistency with 2D frustum in image plane. Attacker runs optimization to move object as far back as possible while retaining at least a minimum IoU (overlap) when projected into 2D image.}
    \label{fig:adv-carla}
\end{figure}

Many systems combine dense 2D image data with sparse 3D point clouds. Prior work showed attacks on 3D data in 2D-3D fusion are stealthy if the attacker retains consistency with unattacked 2D data~\cite{hallyburton2023partial}. We consider an optimal \emph{frustum attack}~\cite{hallyburton2022security} (derived in Appendix~\ref{appendix:threat-optimal}) that exploits the DNN equivariance property to obtain such a stealthy outcome. Fig.~\ref{fig:adv-carla} shows an adversary shifts a car’s 3D bounding box while being stealthy by preserving high overlap with 2D detections. The intersection over union (IoU) threshold determines the maximum translation, enabling displacements over $40~m$ while maintaining IoU $>0.9$. Due to the retained consistency with unaltered 2D detections, single-frame integrity checks fail, making the attack undetectable. Such attacks pose serious risks for path planning, leading to safety incidents.
\section{Neuro-Symbolic AI: A New Paradigm for Perception} \label{sec:nsai}

The widespread effectiveness of attacks on perception underscores the limitations of traditional algorithms in defending against manipulations to the semantic structure of a scene. Thus, motivated to overcome the pattern-matching nature of DNNs, we propose and evaluate a novel neuro-symbolic approach to perception that jointly detects objects and reasons over their semantic relationships.

\subsection{Overview of Approach}

Unlike DNNs, human perception seamlessly integrates low-level feature recognition with high-level reasoning. Humans can infer relationships, identify inconsistencies, and reason about cause-and-effect from vision alone. To safeguard the future of AI-driven autonomy, it is imperative to develop assured perception algorithms that incorporate enhanced contextual awareness and reasoning.

One promising approach to neuro-symbolic perception is scene graph generation (SGG). In addition to detecting objects, SGG yields \emph{inter-object relationships} describing scene composition. Whereas previous attempts to secure perception relied on hardening DNNs to specific attacks, we propose reasoning over semantic scene graphs to evaluate general data integrity. Such graphical models are able to hold high-level semantic insights that 2D object detectors alone fail to capture. In this section, we discuss graph generation, integrity evaluation, and graph-informed sensor fusion.

\begin{figure}
    \centering
    \includegraphics[width=0.96\linewidth]{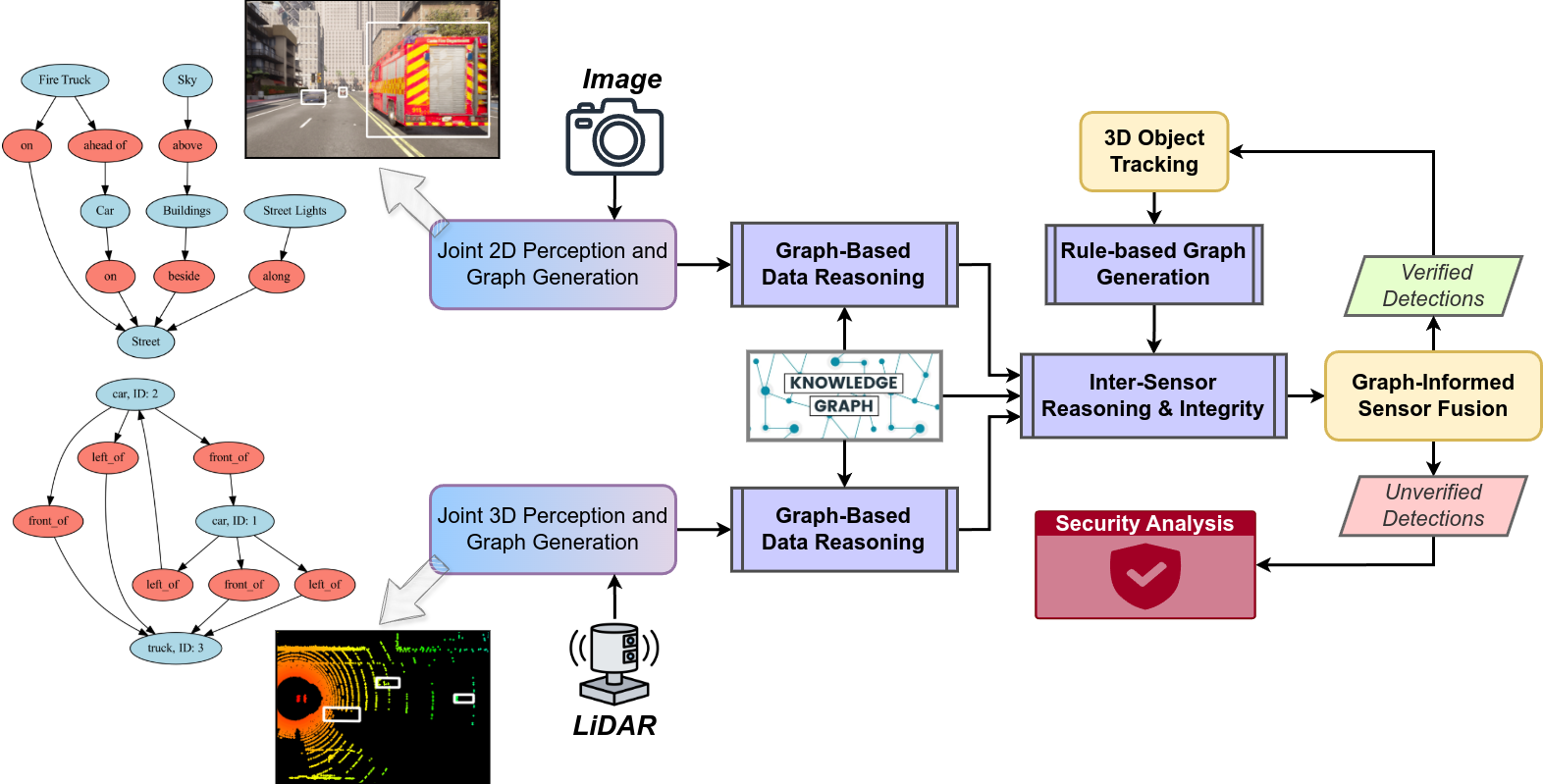}
    \caption{Neuro-symbolic paradigm for perception performs object detection, classification, and scene graph generation jointly, enabling context-based reasoning over e.g.,~spatial relationships from multi-modal data. Reasoning over the graphical models is informed by physics-based knowledge bases and happens both for each sensor and between sensors before impacting sensor fusion.}
    \label{fig:sgg-block-diagram}
\end{figure}

\subsection{System Components}

The proposed neuro-symbolic framework consists of the following components, depicted in Fig.~\ref{fig:sgg-block-diagram}.

\subsubsection{Joint Perception and Graph Generation}

Graphs codify the structure of a scene, and SGG explicitly infers objects and their relationships to yield high-level context-driven scene understanding, often lifting 3D-like relationships (e.g.,~relative positions) directly from 2D data such as images. Early CNN-based SGG approaches struggled to effectively capture relationships due to the inability of CNNs to maintain features between spatially separated objects~\cite{johnson2015image}. However, recent transformer-based methods have demonstrated superior capabilities in modeling interactions across the input space, significantly advancing SGG performance~\cite{carion2020detr}. We describe approaches to SGG below and present examples in Figs.~\ref{fig:building-graphs-rules}, \ref{fig:build-graphs-llm}, and \ref{fig:building-graphs-sgg}. Additional discussion on SGG algorithms is provided in Appendix~\ref{appendix:sgg}. 

\renewcommand{\subfiggraphicswidth}{0.35}
\begin{figure}

    \centering
    \subfigure[Bird's eye view (BEV) of 3D box detections on point cloud.][t]{
        \label{fig:building-graphs-rules-pc}
        \includegraphics[trim={0cm 1.5cm 0cm 1.5cm},clip,width=\subfiggraphicswidth\linewidth]{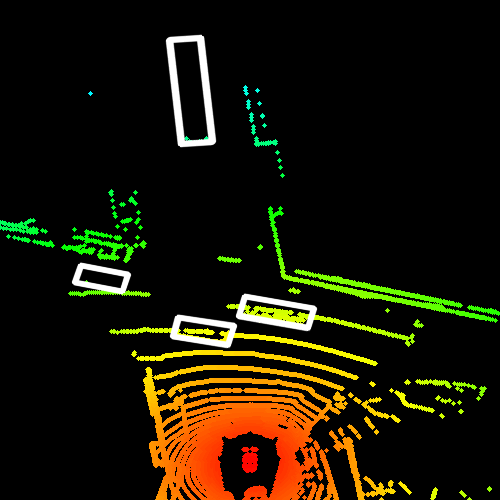}
    }
    \hspace{8pt}
    \subfigure[Rule-based scene graph using as input 3D box detections.][t]{
        \label{fig:building-graphs-rules-graph}
        \includegraphics[trim={0.5cm 0.5cm 0.5cm 0.5cm},clip,width=\subfiggraphicswidth\linewidth]{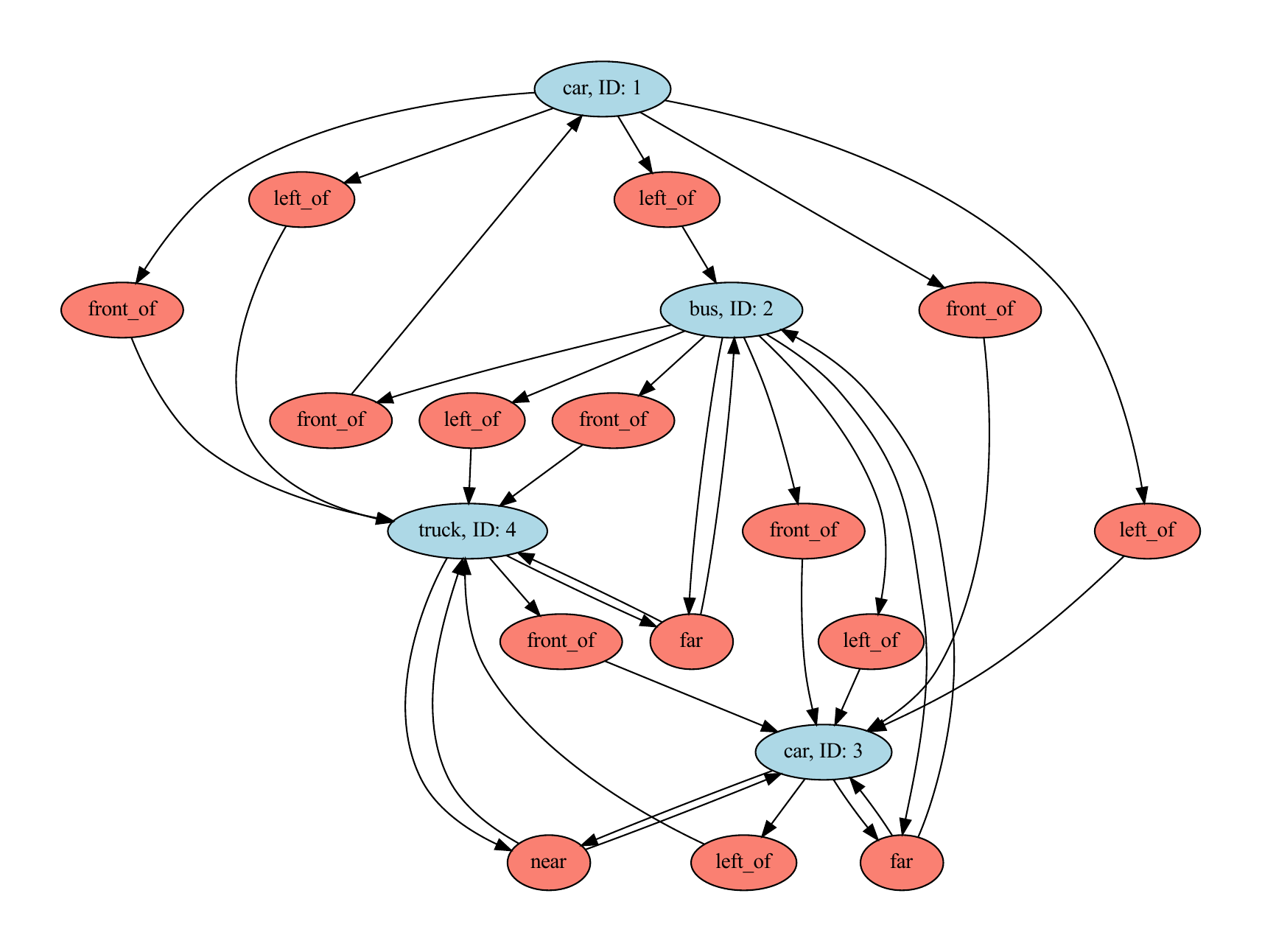}
    }
    \caption{Scene pairs with below images. (a) BEV projection of LiDAR point cloud from nuScenes dataset shown with box detections. (b) Geometric rules build scene graphs using 3D boxes. Nodes (blue) connected via edge relations (red).}
    \label{fig:building-graphs-rules}

    \vspace{8pt}
    \centering
    \subfigure[Raw image used to jointly detect objects and build graph.][t]{
        \label{fig:building-graphs-llm-image}
        \includegraphics[trim={0.5cm 0cm 0cm 0cm},clip,width=\subfiggraphicswidth\linewidth]{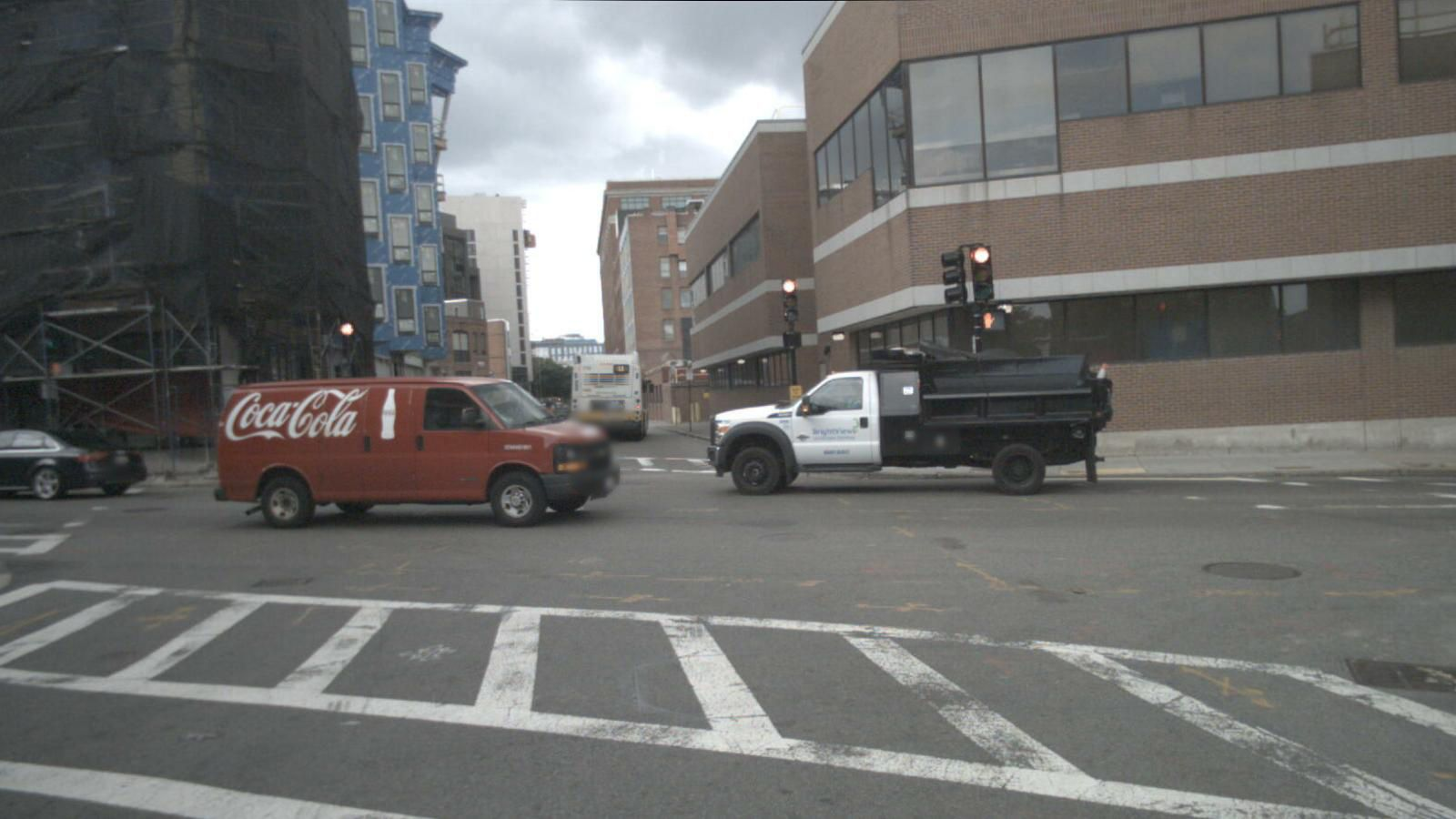}
    }
    \hspace{8pt}
    \subfigure[Foundation model builds a scene graph using raw image as input.][t]{
        \label{fig:building-graphs-llm-graph}
        \includegraphics[trim={0.5cm 0.5cm 0.5cm 0.5cm},clip,width=\subfiggraphicswidth\linewidth,height=11em]{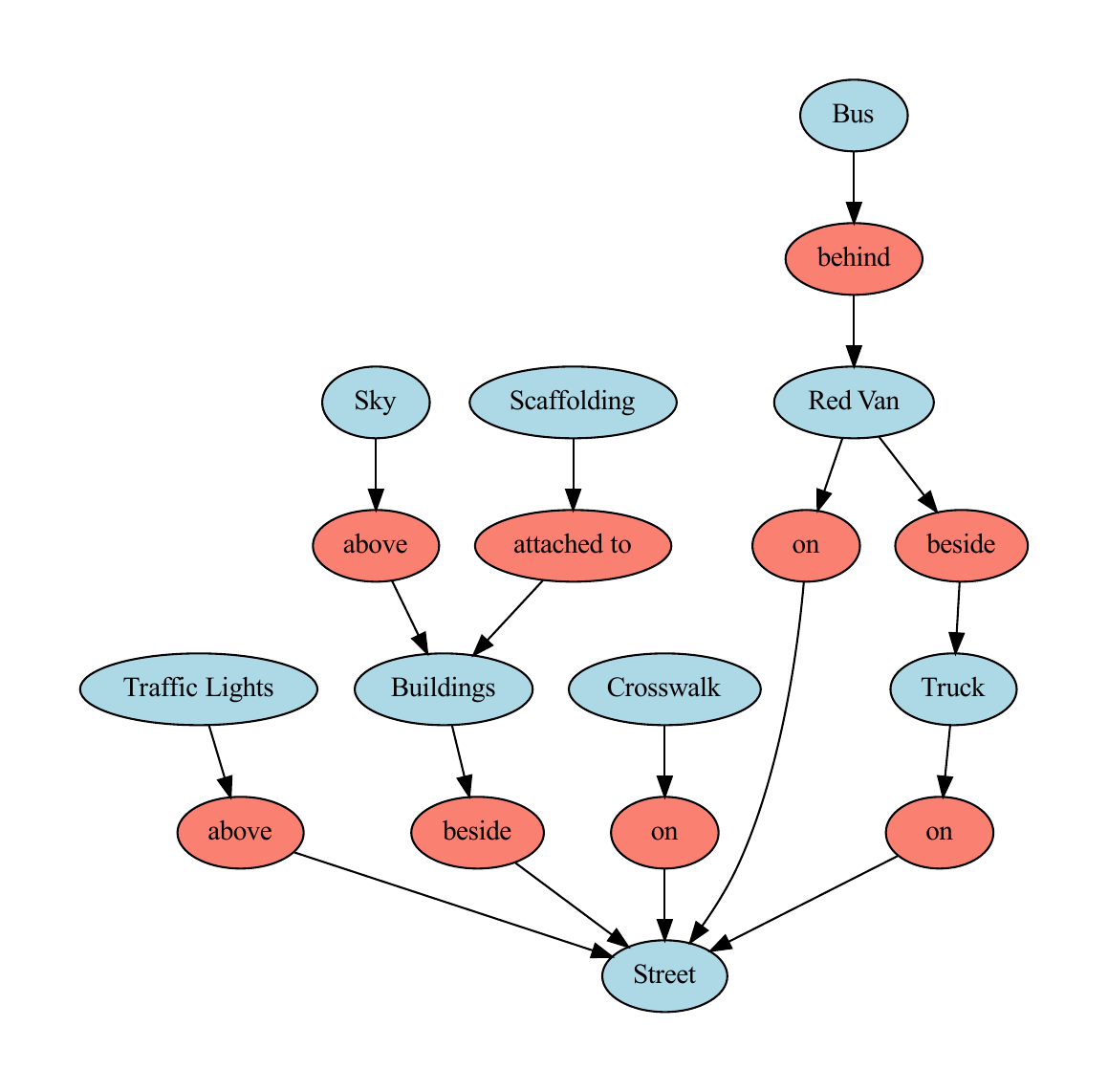}
    }
    \caption{(a) Raw camera image feeds foundation model that (b) directly builds scene graph. Node and edge types differ from rule-based approach because foundation model yields nodes and relationships based on large multi-modal training process.}
    \label{fig:build-graphs-llm}

    \vspace{8pt}
    \subfigure[EGTR specialized SGG model trained to detect nodes and predict edges.][t]{
        \label{fig:building-graphs-sgg-model}
        \includegraphics[width=0.9\linewidth]{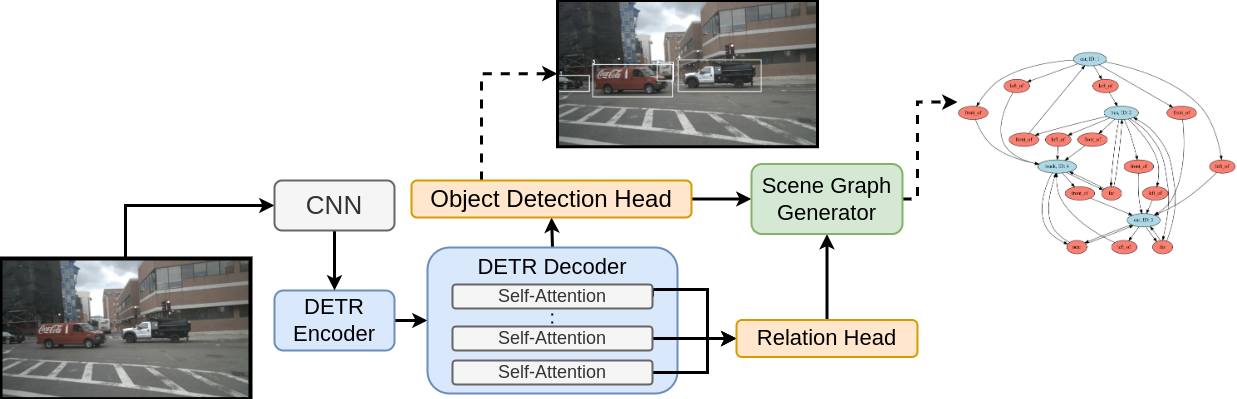}
    }
    \caption{Joint perception and scene graph generation performed using EGTR model from~\cite{im2024egtr}. Regressed graph connects nodes with relations in (subject, predicate, object) format.}
    \label{fig:building-graphs-sgg}

\end{figure}

\paragraph{Geometric Functions (Rule-Based).}
Geometric functions over a 3D Cartesian space define spatial relationships between objects. 3D objects detected from 3D point clouds are passed to manually-defined functions to build relationships such as proximity, adjacency, occlusion, and orientation. This approach is not effective for 2D image data given a camera's lack of explicit distance resolution. Fig.~\ref{fig:building-graphs-rules} and Appendix~\ref{appendix:sgg-rules} describe rule-based SGG from 3D inputs.

\paragraph{Foundation Models.}
Foundation models, including vision-language transformers like CLIP \cite{radford2021learning} and ViLT \cite{kim2021vilt}, facilitate off-the-shelf SGG by leveraging extensive multi-modal knowledge acquired during training. These models naturally bridge visual and linguistic information, effectively capturing relational semantics, global context, and object interactions. Fig.~\ref{fig:build-graphs-llm} demonstrates SGG using a vision-language foundation model on camera images. To the best of our knowledge, foundation models are not yet capable of SGG from point cloud data.

\paragraph{Specialized SGG Models.}
Transformers excel at modeling complex object relationships and global contexts, making them ideal for SGG. Recent specialized models, such as EGTR \cite{im2024egtr} and SGTR \cite{li2022sgtr}, integrate object proposals with relationship prediction heads to generate comprehensive scene graphs. Fig.~\ref{fig:building-graphs-sgg} illustrates the pipeline and output of a specialized SGG model. SGG models are capable of operating on image or point cloud data, if sufficiently trained. 

\subsubsection{Per-Sensor Graph-Based Integrity Reasoning}

By structuring perception inference outcomes into graphs, SGG captures spatial, semantic relationships from real-world sensor data. To evaluate the self-consistency of graph structures to support assured decision-making and safety in autonomy, downstream integrity then reasons over the graph contents. We introduce an approach to reasoning leveraging knowledge graph embeddings (KGEs) and constraint satisfaction evaluation (CSE), both concepts highlighted in~\cite{dimasi2023scene}.

KGEs are structured representations of commonsense information and can enhance the assessment of scene graph integrity,~\cite{chen2020review}. Anomaly detectors compare extensive knowledge from KGEs with outcomes of SGG to validate real-world observed concepts against generated relationships. Integrating knowledge graphs allows models to \emph{infer and correct} potentially erroneous scene interpretations based on well-established semantic relationships. This approach enhances the robustness and integrity of SGG, leading to more accurate representations of scenes.

Constraint satisfaction provides an effective means for assessing the integrity of scene graph outputs by enforcing logical and semantic requirements on the relationships among detected objects. By applying domain-specific constraints, anomaly detectors can either reject unrealistic scene graph outputs or guide corrective measures, ultimately ensuring outputs align more closely with real-world knowledge and expectations, such as the logical requirements outlined in~\cite{giunchiglia2023road}.

\subsubsection{Cross-Sensor Graph-Based Integrity Reasoning}

Classical anomaly detectors such as $\chi^2$ innovation tests on state estimators or inter-sensor assignment between detections from multiple sensors only evaluate consistency at the individual detection level. In contrast, our neuro-symbolic cross-sensor integrity function takes into account the full graph of detections and their relationships with other detections. This approach to cross-sensor integrity is particularly effective in securing perception to attacks that alter the semantic understanding of the scene, such as a false positive/negative or translation attack. Even if graphs are self-consistent, they may not be consistent across sensors; cross-sensor evaluation enhances security robustness if attackers cannot consistently compromise all sensors at once. 

Strategies for cross-sensor graph integrity reasoning include \emph{both brute-force and learned-inference} algorithms. With the brute-force approach, nodes are matched between graphs, and edges traveling between nodes are then matched based on the node matching. Any edges without a match are evaluated to determine if the lack of agreement is a product of noisy sensor data or an attack. A complementary approach is to feed scene graphs to graph neural networks (GNNs) to evaluate for consistency. The result of graph integrity is a per-node, per-edge classification of (in)consistent. 

Beyond anomaly detection, \emph{cross-sensor integrity can be used to hypothesize the exact perturbations responsible for any observed inconsistencies}, such as identifying adversarial manipulations affecting spatial relationships. This deeper reasoning and threat identification enhances responsiveness to complex adversarial scenarios and presents first-of-a-kind resilience in autonomy.

\subsubsection{Graph-Informed Sensor Fusion}

While traditional sensor fusion algorithms update situational awareness with detections and per-object features, providing fusion with relationships from graphs can significantly enhance performance and robustness to challenging natural and adversarial circumstances. Scene graphs illuminate high-level semantics that object detectors alone are incapable of providing. Such information can help align heterogeneous data (e.g.,~resolve conflicts in inter-sensor object assignment) and reduce ambiguity due to e.g., occlusions. Integrating graphical information into fusion facilitates enhanced contextualization of knowledge and ensures consistency across diverse sensor modalities, ultimately improving situational awareness and reliability in perception systems.
\section{Feasibility Study} \label{sec:experiments}

We present a feasibility study demonstrating how neuro-symbolic methods can detect previously stealthy attacks on perception presented in Section~\ref{sec:vulnerability-case}. We employ ground-vehicle datasets from both the physics-based simulator CARLA and the real-world nuScenes dataset. While multiple scenes were analyzed, due to space limitations, here we present detailed results from one representative CARLA scene. Additional analyses and case studies are provided in Appendix~\ref{appendix:experiments}.

\renewcommand{\subfiggraphicswidth}{0.44}
\begin{figure}
    \centering
    \subfigure[Detections and scene graph built from foundation model on camera input.][t]{
        \label{fig:carla-case-1-a}
        \includegraphics[width=\subfiggraphicswidth\linewidth]{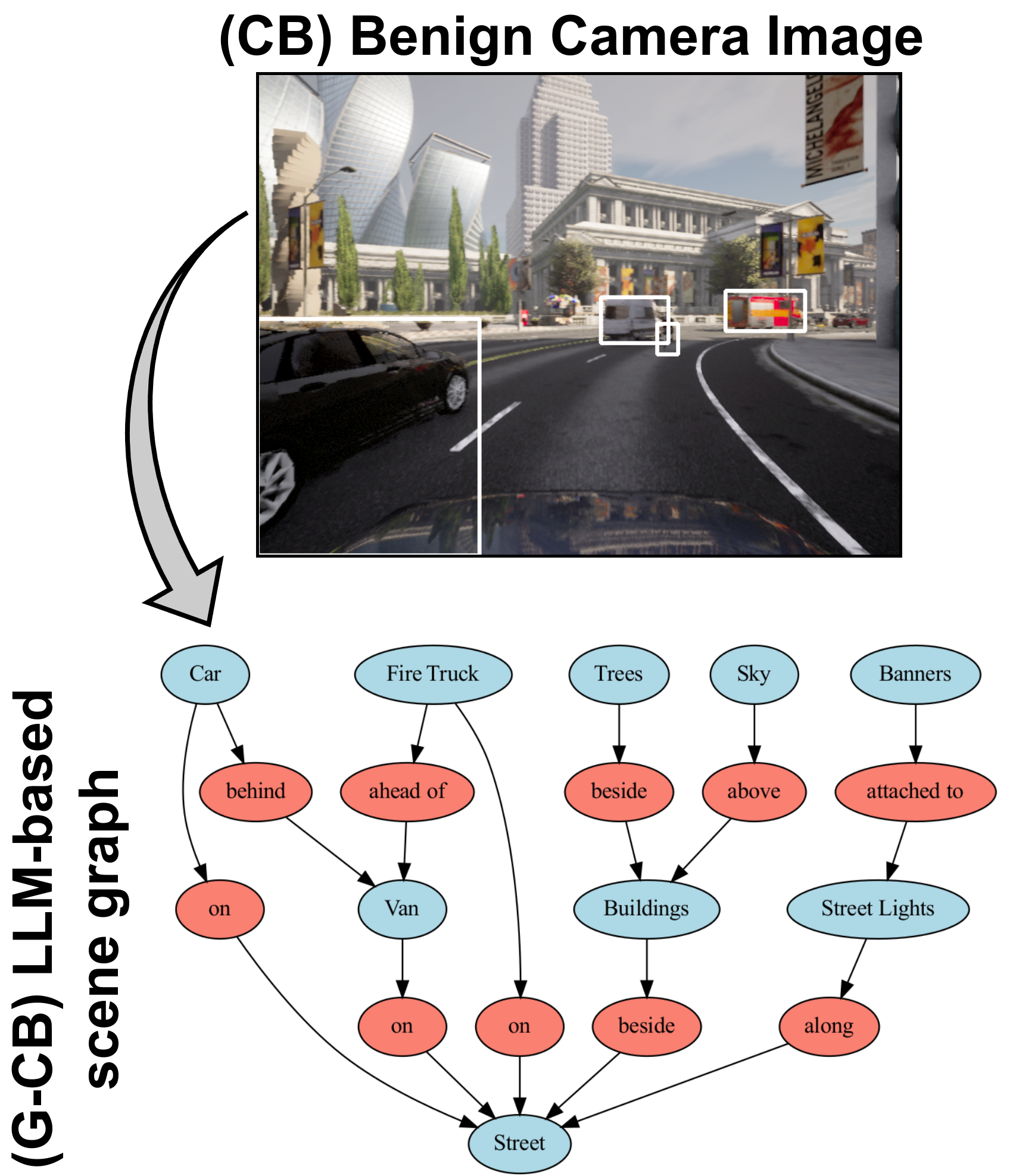}
    }
    \hspace{8pt}
    \subfigure[DNN yields detections on benign LiDAR data, rules construct scene graph.][t]{
        \label{fig:carla-case-1-b}
        \includegraphics[width=\subfiggraphicswidth\linewidth]{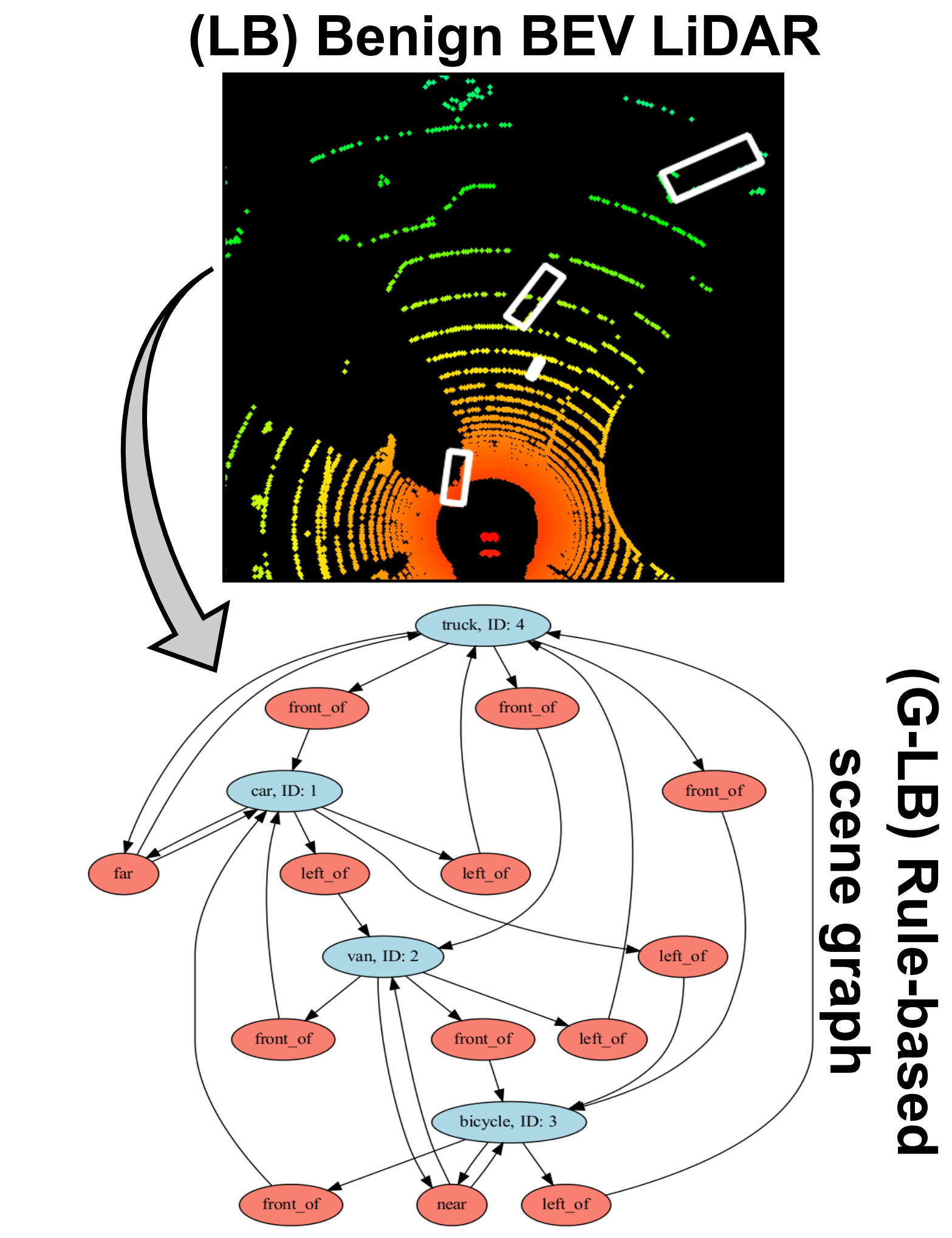}
    }
    \vspace{12pt}
    \subfigure[Adversary manipulates Van, translating it (red box) away from ego. When projected to front-view, Van is still consistent with camera.][t]{
        \label{fig:carla-case-1-c}
        \includegraphics[width=\subfiggraphicswidth\linewidth]{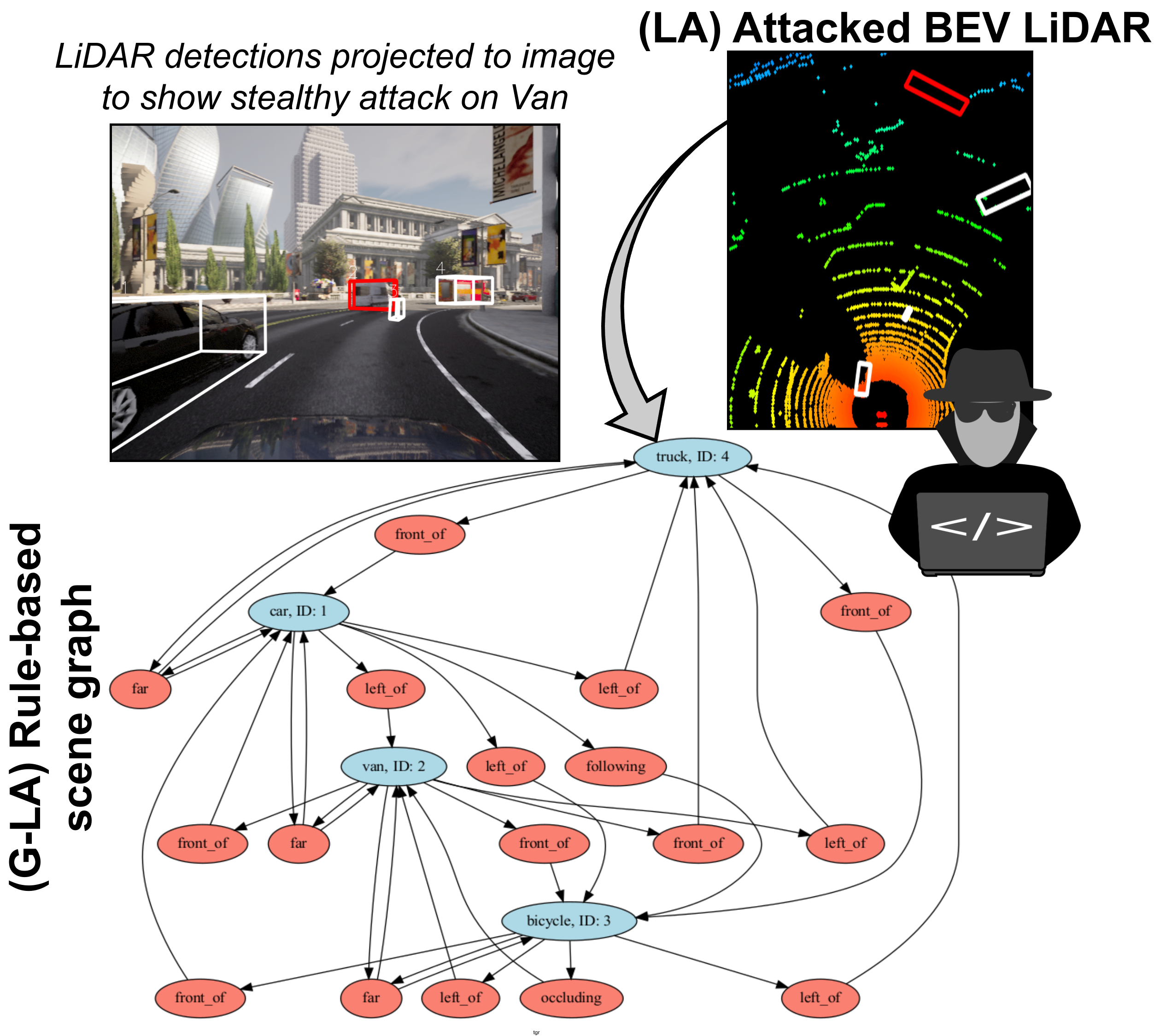}
    }
    \hspace{8pt}
    \subfigure[Reasoning on subgraphs illuminates inconsistencies in semantic concepts between image graph and attacked LiDAR graph.][t]{
        \label{fig:carla-case-1-d}
        \includegraphics[width=\subfiggraphicswidth\linewidth]{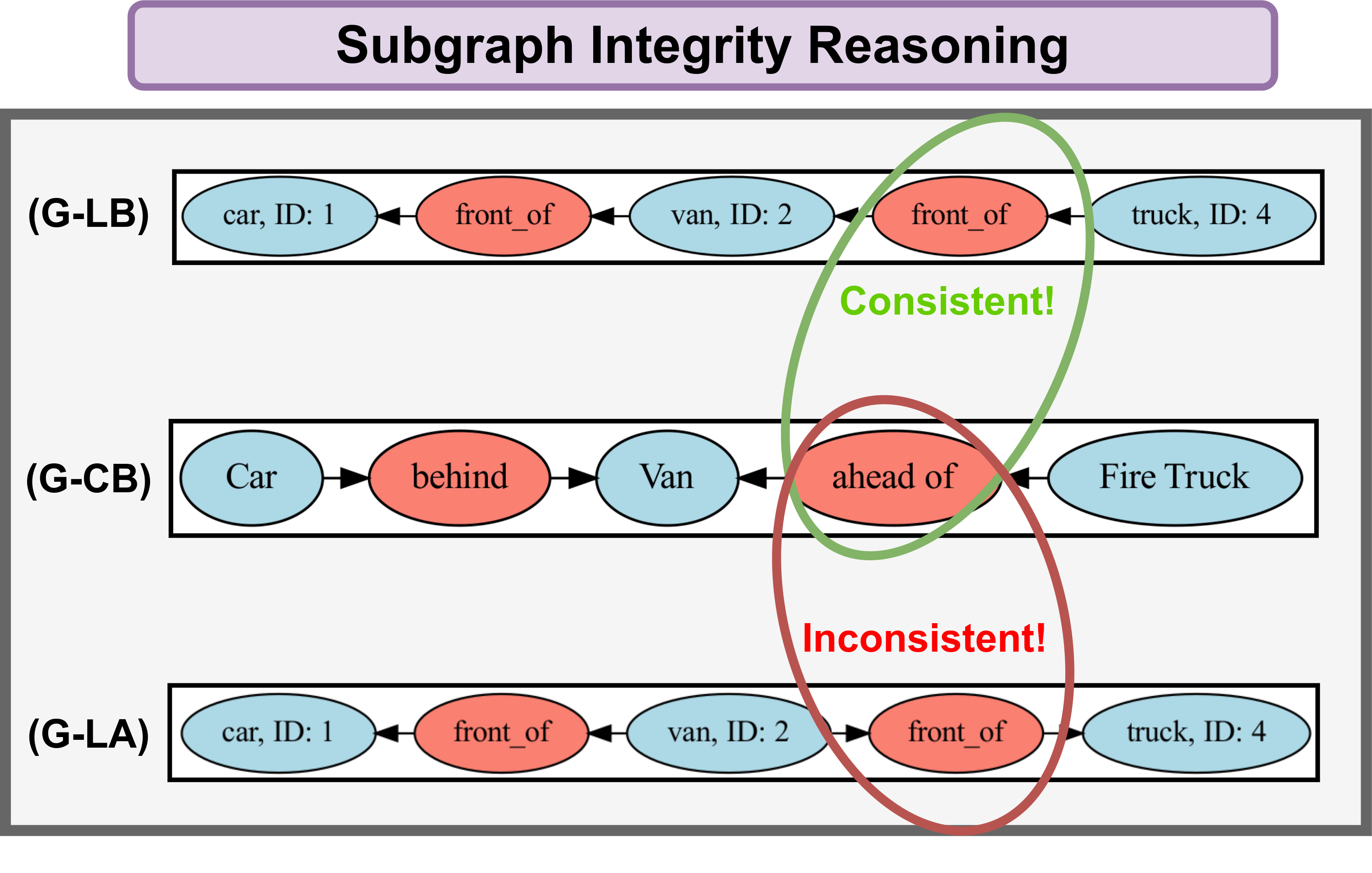}
    }
    
    \caption{(a) Foundation model jointly detects objects and builds scene graph from image. (b,c) Perception yields rule-based scene graph from LiDAR. (c) Attacker translates Van away from the ego - attacked box when projected to camera is still consistent with 2D detections, so camera detections alone cannot detect the attack. (d) Graph-building lifts purely 2D camera data to relational 3D space by inferring positional relationships with context. Inconsistencies identified between camera and LiDAR graphs allow for attack detection of previously thought-to-be stealthy attacks.}
    \label{fig:carla-case-1}
\end{figure}


Figs.~\ref{fig:carla-case-1-a} and~\ref{fig:carla-case-1-b} present benign camera and LiDAR data from a scene that includes four detected objects: a nearby car, a mid-distance van and bicycle, and a distant truck. Traditional camera- and LiDAR-based detectors accurately regress bounding boxes for these objects.

\subsection{Scene Graph Generation}

We employ a foundation model to construct graphs from images through natural language prompts. Despite using 2D images, these models can effectively infer 3D spatial relationships between objects. Each image is analyzed with the query: \texttt{"Build a scene graph from this image."} The foundation model outputs \texttt{(subject, predicate, object)} triplets where \texttt{subject} and \texttt{object} are detected instances in the scene and \texttt{predicate} falls within the set of relationships such as \texttt{in front of}, \texttt{near}, \texttt{occluding}, \texttt{following}. The set of considered relationships for this work is described in Appendix~\ref{appendix:sgg}. An example is illustrated in Fig.~\ref{fig:carla-case-1-a}.

For LiDAR data, scene graphs are generated by first detecting 3D bounding boxes using classical detectors and then passing boxes to rule-based geometric relationship functions. This study focuses on proximal relationships (e.g., \texttt{front of}, \texttt{left of}, \texttt{near/far}). Appendix~\ref{appendix:sgg-rules} describes all considered rules. Fig.~\ref{fig:carla-case-1-b} illustrates a LiDAR-derived scene graph.

\subsection{Adversary Threat Model}

Section~\ref{sec:vulnerability} described stealthy attacks exploiting differences in sensor resolution, such as frustum attacks, where an object is repositioned in 3D space yet maintains consistency with 2D data, thus passing through multi-sensor fusion undetected. Fig.~\ref{fig:carla-case-1-c} demonstrates such an attack where the van's position is significantly translated in LiDAR data. This attack is stealthy to traditional DNN-based detectors due to the retained consistency when 3D detections are projected to the image. 

\subsection{Integrity Evaluation via Scene Graphs}

The scene graph of the 2D image (Fig.\ref{fig:carla-case-1-a}) aligns with the original, unattacked LiDAR-derived graph (Fig.\ref{fig:carla-case-1-b}). In contrast, the graph from compromised LiDAR data (Fig.\ref{fig:carla-case-1-c}) shows clear inconsistencies with the image's graph. Cross-sensor integrity in Fig.\ref{fig:carla-case-1-d} highlights that inconsistencies in the van-truck relationship across sensors are easily identifiable. Neuro-symbolic perception and integrity extracts meaningful semantic information from sensor data, enabling attack detection even though the manipulated 3D box remains consistent with the unattacked 2D boxes.

The neuro-symbolic SGG pipeline offers the first method that secures perception against attacks exploiting asymmetric sensor resolutions, such as the frustum attack. While we use this style of attack as a case study, perception can benefit broadly from neuro-symbolic reasoning. While DNNs detecting objects from images only capture 2D information, SGG lifts relationships from the image that contain 3D insights by leveraging context in the scene. The extraction of relational scene graphs facilitates a more insightful comparison between data and supports secure and assured autonomy. 
\section{Challenges to Realizing Neuro-Symbolic Perception} \label{sec:challenges}

\paragraph{Real-time constraints.} Deploying foundation models (e.g., LLMs, vision transformers) in CPS is challenging due to high computational demands and cloud latency. To address this, we propose training specialized SGG algorithms on autonomy-specific datasets for efficient, real-time SGG.

\paragraph{Dataset construction.} Constructing high-quality datasets while handling edge cases (e.g., zero-/few-shot) and real-world complexity is a key challenge for neuro-symbolic algorithms~\cite{gilpin2021neuro}. We employed a dataset generation pipeline using AVstack~\cite{hallyburton2023avstack} and CARLA from~\cite{hallyburton2023datasets} to being construction of the first neuro-symbolic datasets in autonomy. See Appendix~\ref{appendix:sgg-datasets} for details.
\section{Conclusion and Future Research Directions} \label{sec:conclusion}

To address the lack of security guarantees in existing AI models that operate as simple pattern-matchers, we proposed a paradigm shift toward \textit{neuro-symbolic perception} in safety-critical domains, integrating logical reasoning and commonsense knowledge with deep neural networks for tasks including object detection. By leveraging SGG and foundation models for structured environment understanding, our approach bridges the gap between low-level sensor perception and high-level reasoning. Through feasibility studies in physics-based simulators and real-world datasets, we demonstrated that SGG enhances resilience, interpretability, and security in AI-driven autonomy, paving the way for more trustworthy and robust perception in safety-critical applications.

\section*{Future Research Direction}

Advancing neuro-symbolic sensor fusion requires substantial investment. The following are identified key areas that will drive the maturation of this technology toward deployment in real systems.

\paragraph{Temporal Graph Integrity.} Evaluating the longitudinal consistency of graphs ensures that nodes and their relationships evolve consistent with physics. Inexplicable temporal discontinuities will draw integrity scrutiny. Algorithms in this area can draw insights from object tracking.

\paragraph{Knowledge Graphs.} This work proposed per-sensor integrity that integrates KGEs with SGG. While our case study focused on multi-sensor integrity, future research should develop and implement graph consistency evaluations coupled with KGEs.

\paragraph{Multi-Sensor Integrity.} We used a brute-force evaluation of all subgraphs to detect inconsistencies in multi-sensor reasoning. Unfortunately, uncertain or incomplete graphs due to noisy data can yield inaccurate conclusions. Future effort should be spent designing inference algorithms that are robust to noise and capable of probabilistic reasoning over uncertain graphs.

\paragraph{Specialized SGG.} Our presented case study relied on foundation models which are computationally expensive, memory-intensive, prone to latency, and unbounded in output space. Future efforts will train specialized SGG models with first-of-a-kind autonomy-related neuro-symbolic datasets.

\paragraph{LiDAR-Based SGG.} The current LiDAR-based neuro-symbolic inference pipeline is serialized, first detecting objects before building scene graphs with geometric functions. To ensure consistency with image-based inference and to advance neuro-symbolic capabilities, future research will develop models that directly ingest LiDAR data and jointly perform object detection and SGG.

\acks{This work is sponsored in part by the ONR under agreement N00014-23-1-2206, AFOSR under the award number FA9550-19-1-0169, and by the NSF under NAIAD Award 2332744 as well as the National AI Institute for Edge Computing Leveraging Next Generation Wireless Networks, Grant CNS-2112562.}

\bibliography{references}

\appendix
\section{Data Fusion} \label{appendix:fusion}

One approach to multi-sensor data fusion in autonomy is to obtain detected objects from each sensor in parallel, perform an assignment between the objects detected in pairs of sensors, and fuse the detection data for each assignment set in a state estimator (e.g.,~Kalman filter). We briefly review the general form of \emph{data association} (``assignment problem'') as it applies to multi-sensor data fusion. We denote detected objects from perception as $\Dtwo \leftarrow \texttt{percep}(\text{img})$ and $\Dthree \leftarrow \texttt{percep}(\text{pc})$ where \texttt{percep} are algorithms that take in sensor data and return bounding boxes around objects. 

Given two sets, $\Sone,\, \Stwo$, as well as a weighting function $C:\Sone \times \Stwo \rightarrow \mathbb{R}$, the assignment problem finds a bijection $f:\Sone \rightarrow \Stwo$ such that a cost function 
\begin{equation*}
    \sum_{s\in \Sone} C\left(s,\, f(s)\right)
\end{equation*}
is minimized. Even if the weighting function is nonlinear, the problem is viewed as linear because the cost is a linear sum. In the case of 2D (e.g.,~image) and 3D (e.g.,~LiDAR, radar) detections, sets are $\Sone \coloneqq \Dtwo$ and $\Stwo \coloneqq \Dthree$. With asymmetric sensor resolution (i.e.,~2D/3D), most often $C$ is the intersection over union (IoU) of the 2D/3D bounding boxes in the image plane, i.e.,
\begin{equation*}
    C(d_i^{2D}, d_j^{3D}) = \text{IoU}\left(d_i^{2D},\,\texttt{project}(d_j^{3D})\right),
\end{equation*}
where \texttt{project} is the operation projecting a 3D box to the 2D image plane and $d_i^{2D}, d_j^{3D}$ are individual detections from each set. In practice, the weight function is used to construct a cost matrix enumerating over all pairs of detected objects in 2D and 3D as
\begin{equation*}
    A[i,j] \leftarrow C(d_i^{2D},\, d_j^{3D}),
\end{equation*}
and a linear sum optimizer is then applied to $A$ to yield the optimal bipartite solution to the assignment problem. Candidate assignment algorithms include the Hungarian and Jonker-Volgenant varieties. A threshold is often used so that low affinity pairs (e.g.,~small overlap) are not accepted.
\section{Adversary Framework and Optimal Frustum Attack} \label{appendix:threat}

While prior works considered \( L_p \) norm perturbations, we instead consider a general attacker objective. This objective allows modeling physically-realizable attacks such as the introduction of a false object while abstracting the implementation details (e.g.,~spoofing vs. cyber-attack vs. backdoor).

\subsection{Attacker Knowledge \& Capability}

We consider a generic adversary that has knowledge of existing objects in a scene. The attacker is also able to manipulate any component of existing bounding box detections. A 2D bounding box is a tuple of location \((u, v)\) and box size \((h, w)\). A 3D bounding box is a tuple of position \((x, y, z)\), box size \((h, w, l)\), and orientation \(\theta\). Attacks can be realized by methods that include sensor spoofing attacks formalized in~\cite{cao2019adversarial}, physical adversarial objects such as from~\cite{tu2020physically}, and cyber-based attacks that exploit pipeline vulnerabilities including Trojans in~\cite{hallyburton2023partial, petit2014potential}. For example, frustum-type attacks lead to translations of existing objects as illustrated in~\cite{hallyburton2022security}.

\subsection{Attacker Goal: Optimal Frustum Attack}
One particular attack objective is to move the 3D bounding box detections as far as possible from their original locations while retaining the same assignment pairs as in the unattacked case so as to remain stealthy to detection from any uncompromised 2D data from e.g.,~image-based detections.

\subsection{Practical Constraints} 
Despite an attacker able to manipulate object detections, certain sensible guidelines must be in place to prevent attacks from being easily detectable. These include:

\vspace{-4pt}
\begin{itemize}
    \setlength\itemsep{-4pt}
    \item \textbf{Box volume.} Manipulated object bounding box volume is to be bounded between a $[V_{\text{min}},\, V_{\text{max}}]$ with bounds set from plausible real-world scenarios; e.g., a semi-truck volume is $150~m^3$.
    \item \textbf{Box dimensions.} Any individual dimension of a box is to be bounded consistent with observed data between $\left[\left(h_{\text{min}},w_{\text{min}},l_{\text{min}}\right),\, \left(h_{\text{max}},w_{\text{max}},l_{\text{max}}\right)\right]$.
    \item \textbf{Vertical position.} When tracking ground vehicles, the objects of interest must be on the ground. Thus the attacker should not significantly manipulate the vertical position.
    \item \textbf{Orientation.} Ground vehicles are constrained to be coplanar with the ground. Thus, only the yaw angle, $\theta$, is to be manipulated while pitch and roll are fixed.
\end{itemize}

\subsection{Optimizing the Frustum Attack} \label{appendix:threat-optimal}

Many deployed systems combine dense 2D image data with sparse 3D point clouds. Prior work~\cite{hallyburton2023partial} demonstrated that attacks on such 2D-3D fusion are particularly devastating because intelligent attacks on 3D data can hide in a stealthy null space in the unattacked 2D data. A powerful attacker can execute an optimal frustum attack as the solution to
\begin{equation} \label{eq:relaxed-objective}
    \begin{gathered}
        \Dthreebar = \argmax_{\Dthree} \sum_{p_j \in \Dthree} \norm{\Bar{p}_j^{3D} - p_j^{3D}}  \\
        \text{s.t.,} \quad \text{IoU}\left(d_i^{2D}, \,\texttt{project}(\Bar{d}_j^{3D})\right) \geq \zeta_{\text{min}} \\
        \text{and} \quad V_{\text{min}} \leq \Bar{h}_j \times \Bar{l}_j \times \Bar{w}_j \leq V_{\text{max}},
    \end{gathered}
\end{equation}
where $p_j$ is the position of detection $d_j$, $\Bar{p_j}$ and $\Bar{d_j}$ the attacker-manipulated position/detection, $\zeta_{\text{min}}$ a minimum intersection over union (IoU) threshold set to maintain consistency between the attacked 3D detections and the unaltered 2D data (for stealthiness), and $\texttt{project}$ the operation projecting 3D data to the 2D image plane. 
\section{Scene Graph Generation} \label{appendix:sgg}

\subsection{Shortcoming of CNNs for SGG}

Scene graph generation (SGG) requires identifying objects, attributes, and relationships within an image. While CNNs excel at visual feature extraction, they struggle with relational reasoning, spatial precision, and contextual ambiguity. CNNs capture local features but fail to model global relationships critical for SGG, necessitating additional mechanisms like graph neural networks, attention, or transformers. Pooling operations further degrade spatial resolution, making precise object relationships difficult to define. Moreover, CNNs lack the ability to resolve visual ambiguities, limiting their effectiveness in complex relational reasoning. These limitations highlight the need for alternative architectures to enhance SGG performance~\cite{johnson2015image, lu2016visual}.

\subsection{Modern Approaches for SGG}

While CNNs serve as a foundational step for object detection and feature extraction, effective scene graph generation (SGG) necessitates additional relational modeling techniques. Graph Neural Networks (GNNs) and Transformers have emerged as leading approaches due to their ability to capture relational structures and contextual dependencies~\cite{zellers2018neural, tang2020unbiased}.  

GNNs represent objects as graph nodes and relationships as edges, enabling message passing that propagates semantic and spatial information throughout the graph for enhanced relational reasoning. Transformers utilize self-attention mechanisms to dynamically model global relationships and contextual interactions without requiring explicit graph construction. By integrating GNNs' structured relational representation with Transformers' flexible contextual modeling, recent methods achieve superior accuracy and generalization in SGG~\cite{im2024egtr,li2022sgtr}. This hybrid approach effectively captures complex object interactions that CNN-based architectures struggle to model, advancing the state-of-the-art in scene graph generation.

\subsection{Hand-Coded Rule-Based Scene Graphs} \label{appendix:sgg-rules}

To build scene graphs from 3D bounding box detections, we define spatially-oriented relation functions. Ultimately, these functions aid in building datasets and training SGG inference algorithms. Each of the functions ingests two objects, $O_1$ (subject) and $O_2$ (object), such that a function e.g.,~$\texttt{front of}(O_1, O_2)$ would test that ``$O_1$ is in front of $O_2$''. ``Symmetric'' relations are those whereby there exists a complement relation such that, e.g., $\texttt{front of}(O_1, O_2) \iff \texttt{behind}(O_2, O_1)$. Note that the complement does not strictly need to be included in the graph because it is implied by the former. Therefore, we call a graph ``reduced'' if it contains only one of any complement pairs. The implementations are in the source code that will be released online.

\renewcommand{\itembox}{2.2cm}
\begin{itemize}
    \setlength\itemsep{-4pt}
    \item{\makebox[\itembox][l]{\texttt{front of}} (complement: \texttt{behind})}
    \item{\makebox[\itembox][l]{\texttt{left of}} (complement: \texttt{right of})}
    \item{\makebox[\itembox][l]{\texttt{occluding}} (complement: \texttt{occluded by})}
    \item{\makebox[\itembox][l]{\texttt{following}} (complement: \texttt{followed by})}
    \item{\makebox[\itembox][l]{\texttt{far from}} (complement: self)}
    \item{\makebox[\itembox][l]{\texttt{close to}} (complement: self)}
    \item{\makebox[\itembox][l]{\texttt{next to}} (complement: self)}
\end{itemize}

\subsection{Building Neuro-Symbolic Datasets} \label{appendix:sgg-datasets}

We utilized the CARLA~\cite{dosovitskiy2017carla} and nuScenes~\cite{caesar2020nuscenes} datasets, along with geometric functions above, to construct the first neuro-symbolic dataset for scene graph generation. CARLA, a high-fidelity autonomous driving simulator, provided synthetic yet realistic urban driving scenarios, while nuScenes offered large-scale real-world driving data with detailed 3D annotations. By leveraging these datasets, we extracted object-centric representations, capturing spatial, semantic, and kinematic properties of dynamic and static elements within the scene. Using geometric functions, we computed precise spatial relationships such as distances, occlusions, and patterns between objects, ensuring an explicit and structured encoding of scene interactions. This integration of real-world and simulated data, combined with formal geometric reasoning, enabled the creation of a neuro-symbolic dataset that bridges visual perception with structured relational reasoning, setting a foundation for robust scene graph generation for future research in autonomy.

\begin{figure}[H]
    \centering
    \subfigure[Detections, scene graph built from foundation model on camera input.][t]{
        \label{fig:nuscenes-case-0-a}
        \includegraphics[width=0.4\linewidth]{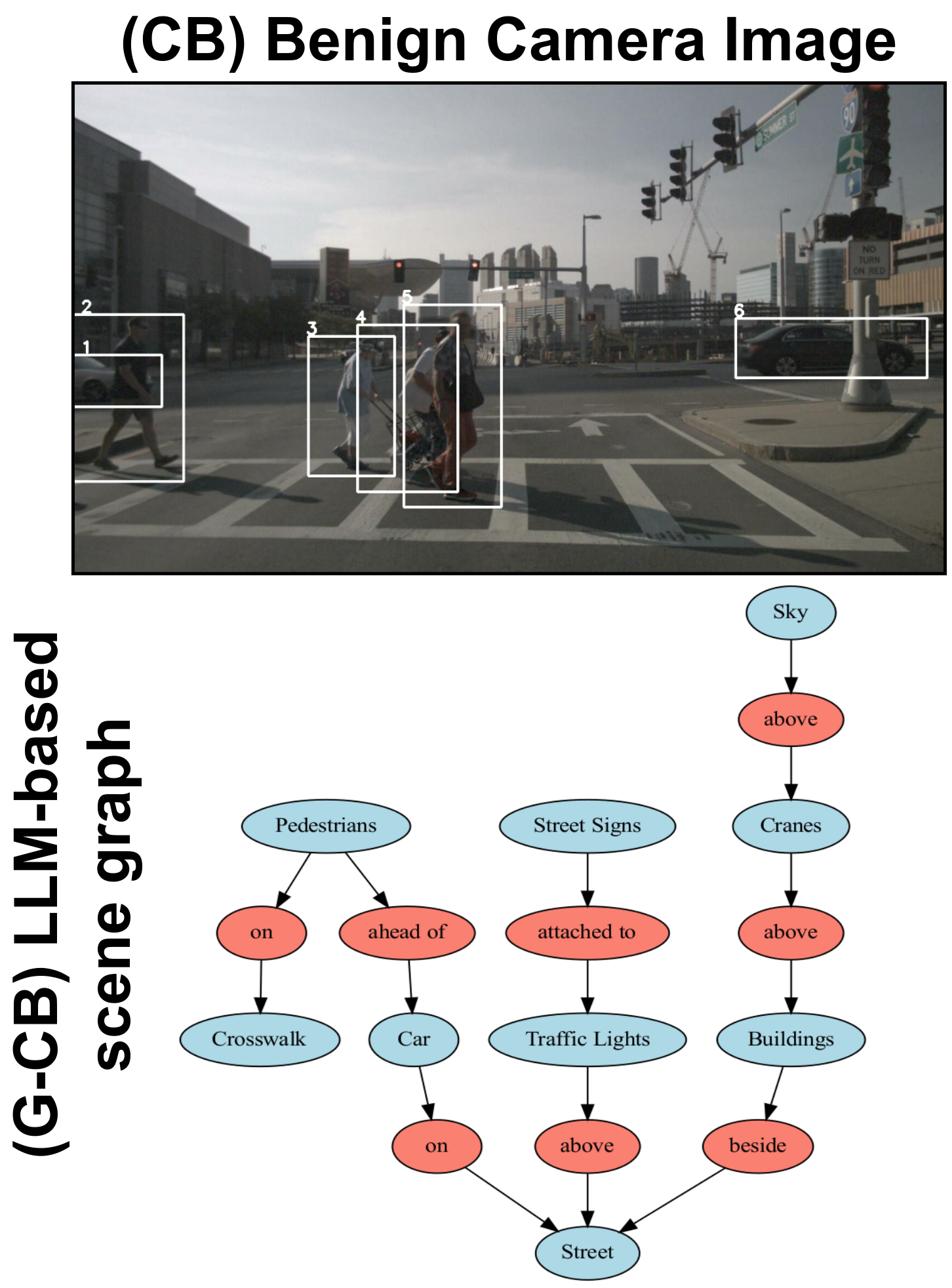}
    }
    \hspace{8pt}
    \subfigure[DNN yields detections on LiDAR data, rules construct scene graph.][t]{
        \label{fig:nuscenes-case-0-b}
        \includegraphics[width=0.4\linewidth]{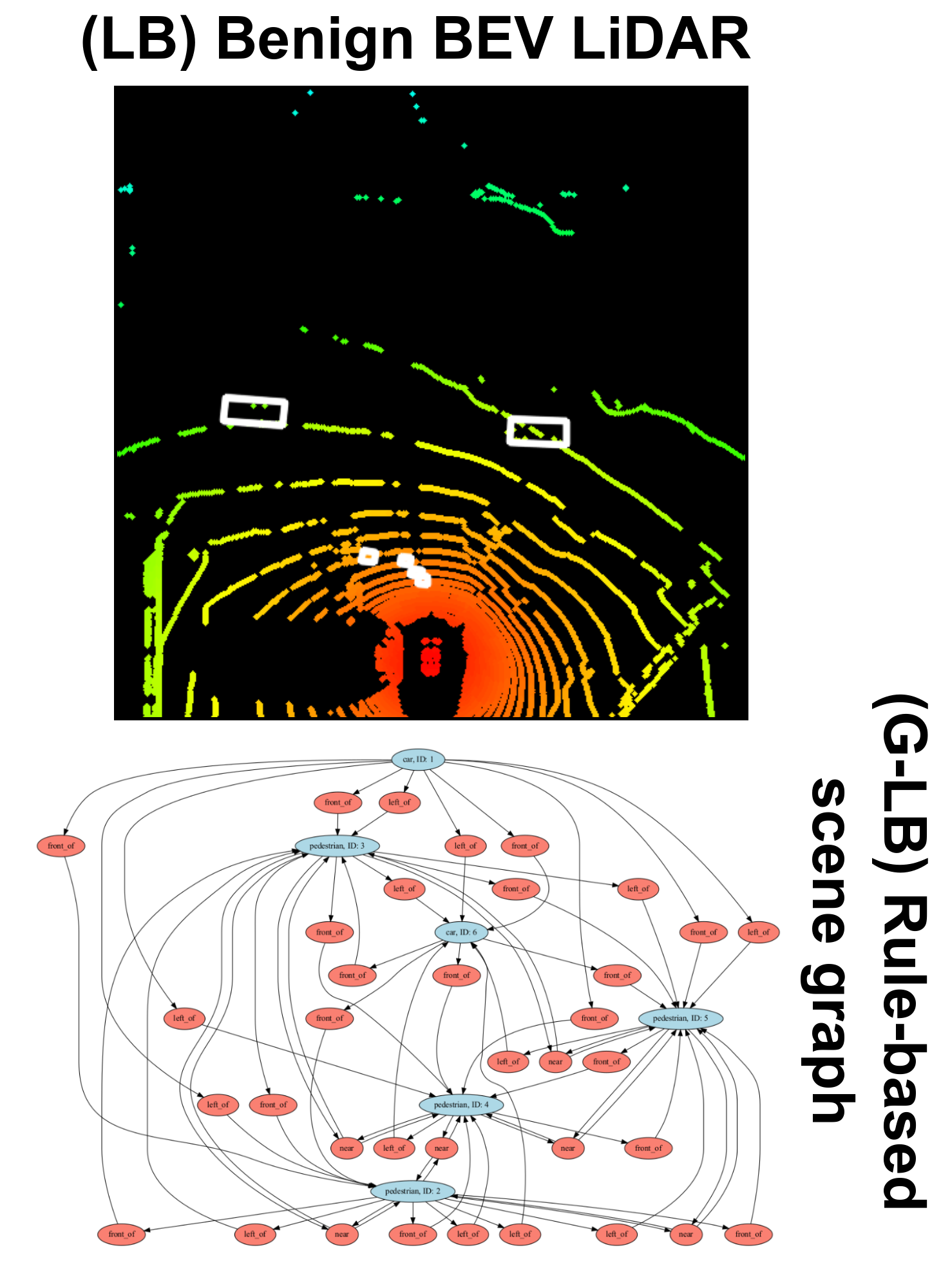}
    }
    \vspace{12pt}
    \subfigure[Adversary manipulates pedestrian, translating it away from ego. When projected to front-view, pedestrian is still consistent with camera.][t]{
        \label{fig:nuscenes-case-0-c}
        \includegraphics[width=0.8\linewidth]{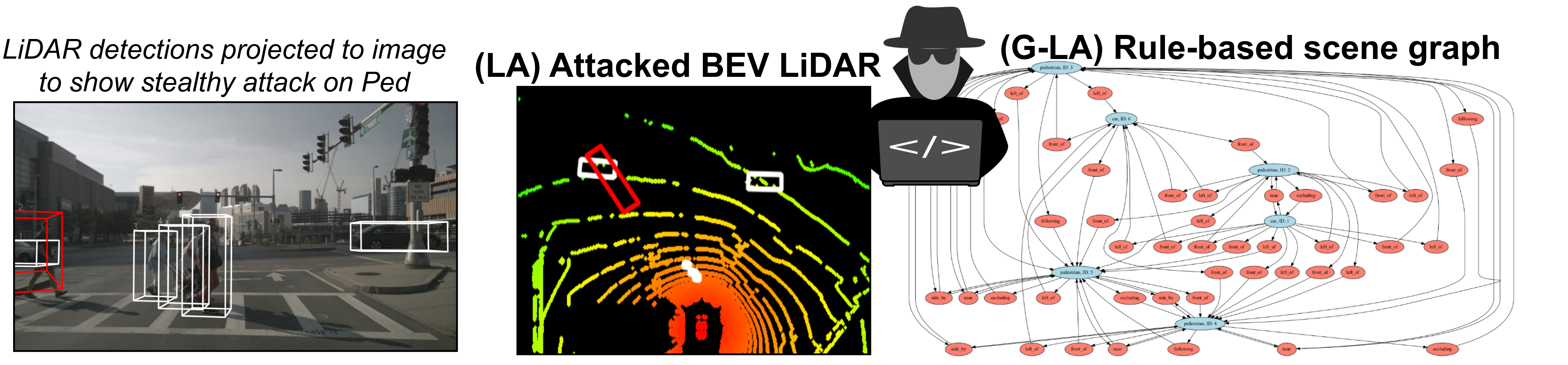}
    }
    \hspace{8pt}
    \subfigure[Reasoning on subgraphs illuminates inconsistencies in semantics between image and attacked LiDAR graph.][t]{
        \label{fig:nuscenes-case-0-d}
        \includegraphics[width=0.7\linewidth]{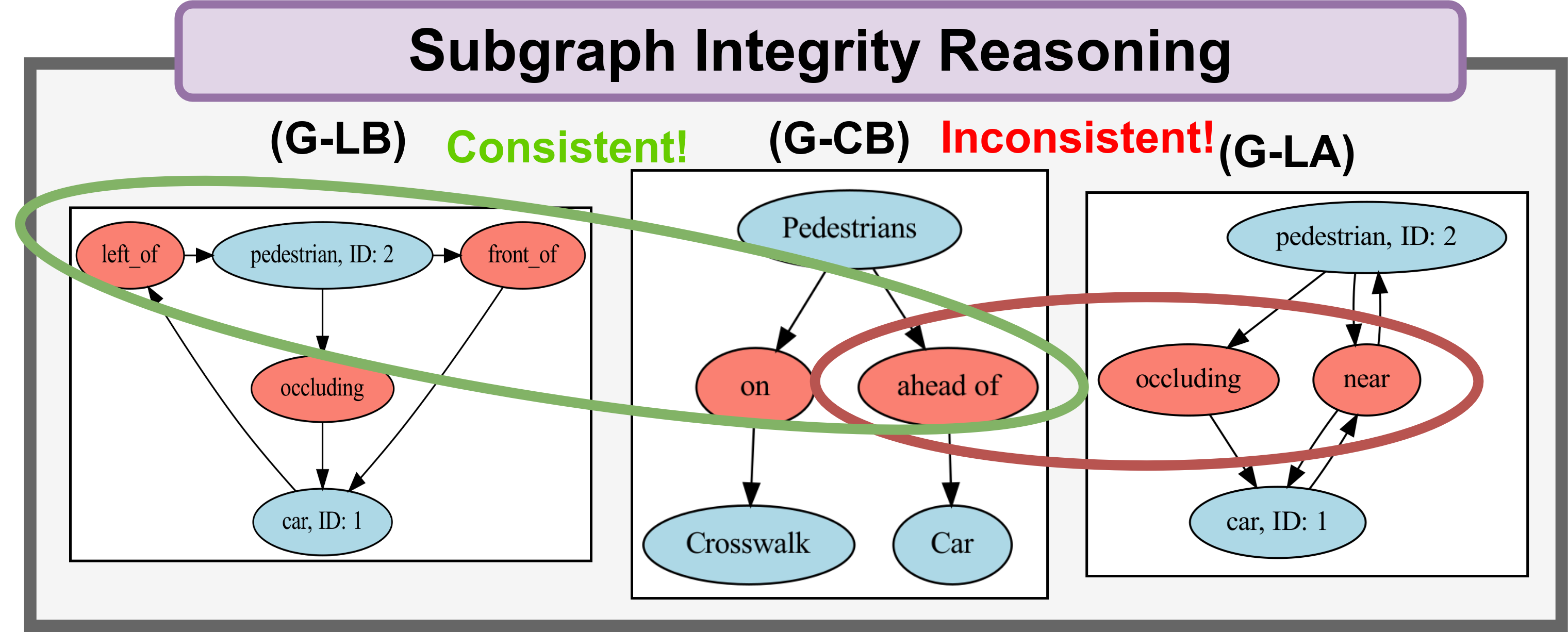}
    }
    
    \caption{Case study of using scene graph generation to secure multi-sensor fusion from attacks on sensing. Analysis procedure follows that of Figure~\ref{fig:carla-case-1}.}
    \label{fig:nuscenes-case-0}
\end{figure}

\begin{figure}[H]
    \centering
    \subfigure[Detections, scene graph built from foundation model on camera input.][t]{
        \label{fig:nuscenes-case-1-a}
        \includegraphics[width=0.4\linewidth]{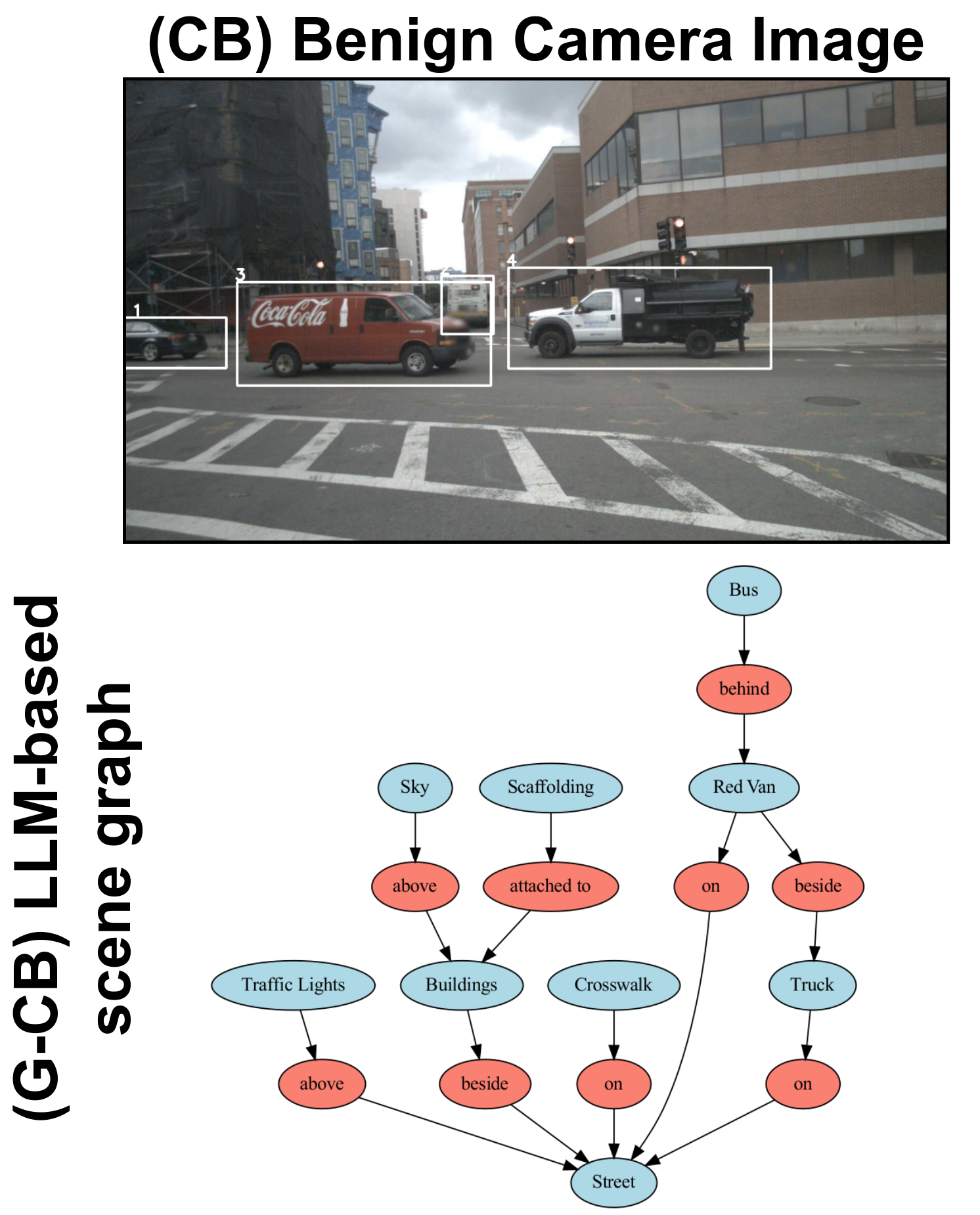}
    }
    \hspace{8pt}
    \subfigure[DNN yields detections on LiDAR data, rules construct scene graph.][t]{
        \label{fig:nuscenes-case-1-b}
        \includegraphics[width=0.4\linewidth]{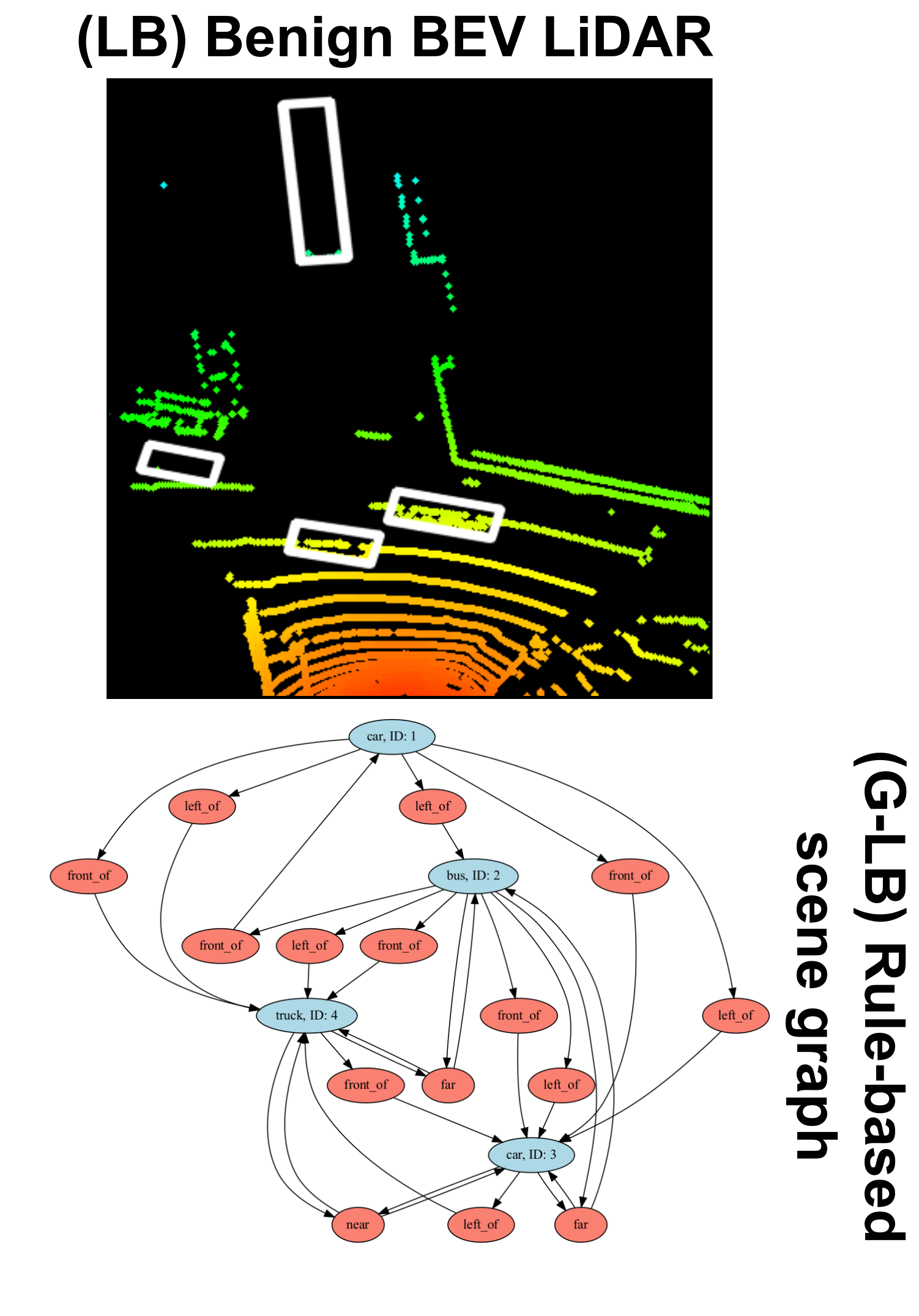}
    }
    \vspace{12pt}
    \subfigure[Adversary manipulates pedestrian, translating it away from ego. When projected to front-view, pedestrian is still consistent with camera.][t]{
        \label{fig:nuscenes-case-1-c}
        \includegraphics[width=0.8\linewidth]{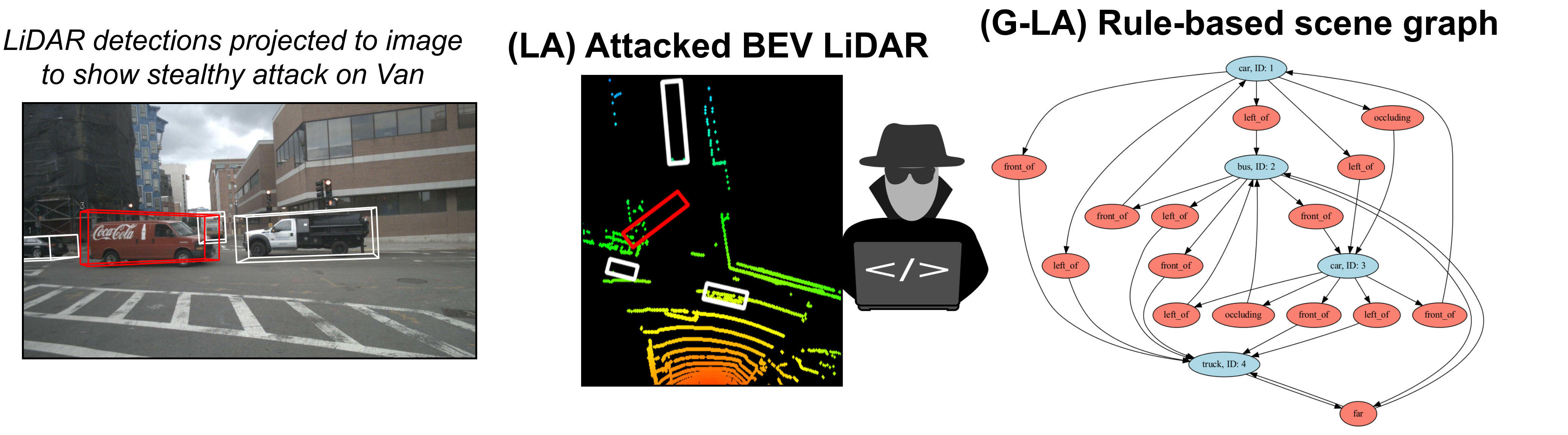}
    }
    \hspace{8pt}
    \subfigure[Reasoning on subgraphs illuminates inconsistencies in semantics between image and attacked LiDAR graph.][t]{
        \label{fig:nuscenes-case-1-d}
        \includegraphics[width=0.45\linewidth]{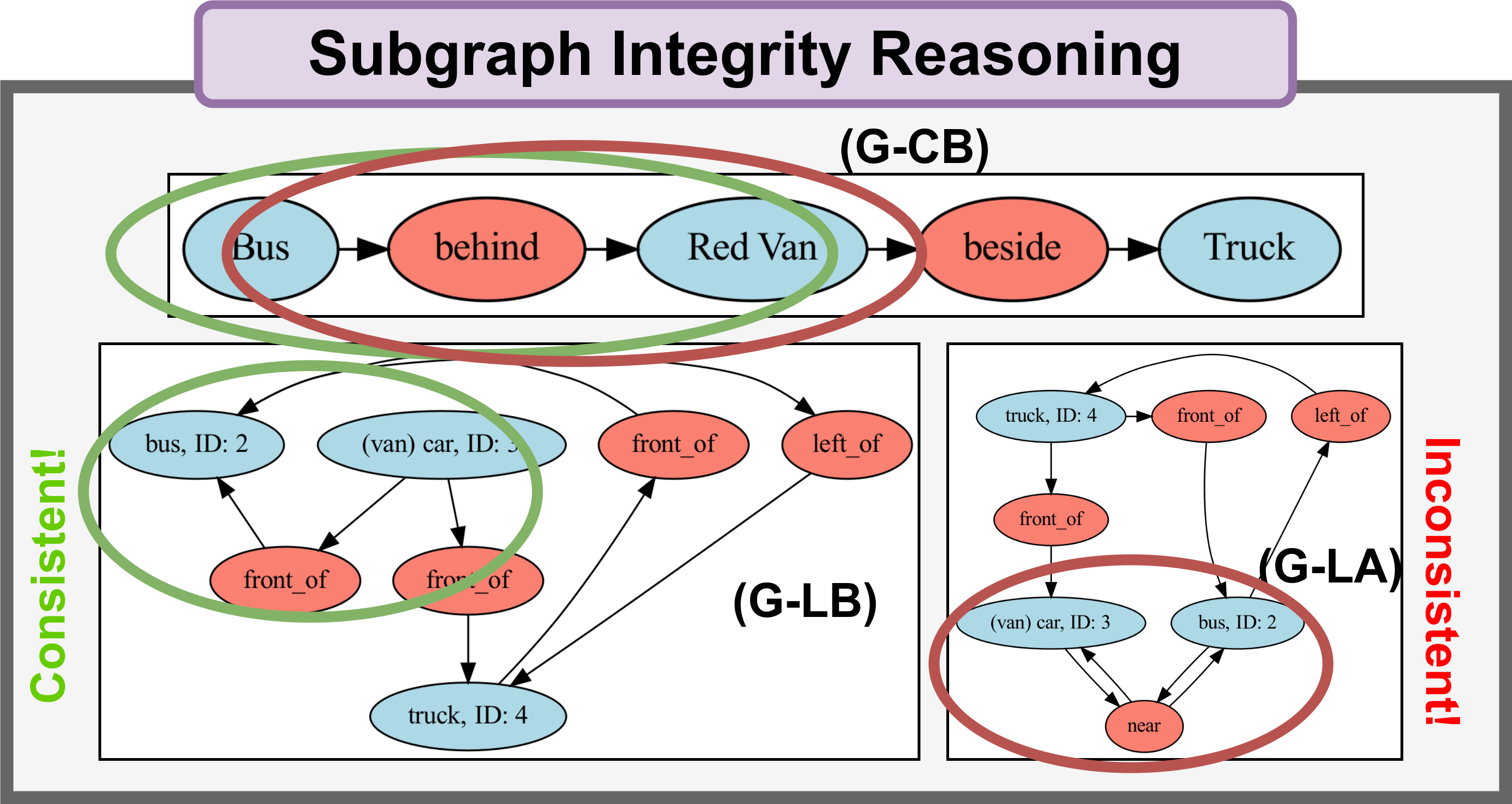}
    }
    
    \caption{Case study of using scene graph generation to secure multi-sensor fusion from attacks on sensing. Analysis procedure follows that of Figure~\ref{fig:carla-case-1}.}
    \label{fig:nuscenes-case-1}
\end{figure}


\section{Supplemental Case Studies} \label{appendix:experiments}

A case study walking through scene graph generation and integrity reasoning from the CARLA simulator was presented in Figure~\ref{fig:carla-case-1}. We illustrate a case where an attacker translates the detection of a pedestrian in a scene from the nuScenes dataset in Figure~\ref{fig:nuscenes-case-0}. Similarly, evaluating the graphs through an integrity function illuminates inconsistencies between the semantics in the perception results. Finally, Figure~\ref{fig:nuscenes-case-1} describes another scene from nuScenes where a van is adversarially translated away from the ego vehicle. Scene graph generation and graph-based integrity is able to detect inconsistencies in the inference results.

\end{document}